# Enhancing Tea Leaf Disease Recognition with Attention Mechanisms and Grad-CAM Visualization

[1]Omar Faruq Shikdar, [1]Fahad Ahammed, [1]B. M. Shahria Alam*, [1]Golam Kibria, [1]Tawhidur Rahman, [1]Nishat Tasnim Niloy

[1]Department of Computer Science and Engineering, East West University, Dhaka, Bangladesh

Email: omorfaruk549@gmail.com, fahadahbd@gmail.com, bmshahria@gmail.com, glkibria4535@gmail.com, tawhidur7@gmail.com, nishat.niloy@ewubd.edu

**Abstract** Tea is among the most widely consumed drinks globally. Tea production is a key industry for many countries. One of the main challenges of tree harvest is tea leaf diseases. If the spread tea leaf diseases are not stopped in time, it can lead to massive economic losses for the farmers. So, it's crucial to identify tea disease as soon as possible. Manually identifying tea leaf disease is an ineffective and time-consuming method, without any guarantee for success. Automating this process will improve both the efficiency and the success rate of identifying tea leaf diseases. The purpose of this study is to create an automated system that can classify different kind of tea leave diseases allowing the farmers to take actions to minimize the damage. A novel dataset was developed specifically for the purposes of this study. The dataset contained 5278 images for seven class. The dataset was pre-processed prior training the model. We deployed three pretrained model, DenseNet, Inception and EfficientNet. EfficientNet was only used in the Ensemble model. We utilized two different attention modules to improve model performance. The Ensemble model achieved the highest accuracy of 85.68%. Explainable AI was introduced for better model interpretability.

## 1. Introduction

Tea is a refreshing beverage that is consumed by the people of Bangladesh and many other countries all over the world. One of the major good that Bangladesh exports is tea [1]. Tea export has a big impact on the country's economy [2]. One of the main challenges of tea harvest is diseases. Different types of fungal diseases are very prominent in tea leaves that yields extremely bad harvest [3]. The occurrence of tea leaf diseases can differ across various regions of the world. Tea algal leaf spot, Brown Blight, Gray Blight, Helopeltis, Red spider, and Green mirid bug are the diseases that occur the most in Bangladesh. Hence, these diseases were incorporated in the dataset which was constructed for this study. In order to stop the spread the diseases and take counter measures in time, it is essential to detect these diseases as early as possible. Artificial intelligence can be a great help in this regard, automating this process will make the whole thing a lot more efficient.

Machine Learning and deep learning both have proven to be quite effective for tea leaf disease classification from images [4]. Especially transfer learning is an efficient technique where pre-trained model is utilized by fine tuning the models for the specific dataset. DenseNet 201 and Inception V3 is two of the most successful pre-trained CNN models. Applying ensemble technique can result in even better model performance. In ensemble technique various models are combined together to create a new model. In this study, a novel dataset was developed, then pre-processed. Then three pre-trained models were deployed, with EfficientNet was exclusively employed for ensemble model. A custom CNN model was placed on top of them. In between them an attention module was placed. Then using various evaluation metrics, the best model was selected. Finally, an explainable AI method called Grad-CAM was integrated with the best model.

## 2. Literature Review

In this section, recent studies that have been done in this area. Numerous deep learning-oriented solutions have been put forward in recent years for the classification and detection of tea leaf diseases, focusing on enhancing the precision, speed, and generalizability of automated systems. Sivaraman et al. explored the application of the VGG16 model in their work on sustainable agriculture, achieving an accuracy of 92% across eight classes of tea leaf diseases. Their approach emphasized the integration of deep learning into crop management systems, providing foundational work for scalable disease detection models [5]. Hairah et al. on the other hand utilized a standard Convolutional Neural Network (CNN) for disease classification, reporting a higher accuracy of 94.55% on a dataset comprising six classes. Their methodology highlighted the effectiveness of traditional CNNs in capturing discriminative features of tea leaves under varied conditions [6]. An advanced model combining CNN with Support Vector Machines (CNN-SVM) was proposed by Kaur et al., which yielded a notable accuracy of 95.01%. Their hybrid architecture demonstrated improved classification performance and robustness, particularly when dealing with overlapping disease symptoms [7]. Conversely, Rahman et al. designed a CNN-based automated detection system specifically for tea leaf diseases in Bangladesh. Their model achieved the highest accuracy of 96.65% across four disease classes, underscoring the potential of regionally trained models to outperform more generalized ones in specific agricultural contexts [8]. Heng et al. proposed an innovative deep neural network method that leverages hybrid pooling layers to optimize feature extraction. Their model attained an accuracy of 92.47% across seven classes and contributed to ongoing research on pooling strategies within deep architectures [9]. Kaur et al. also presented a CNN model, but with relatively lower performance, achieving 82% accuracy across eight classes. This result indicates a possible limitation in training strategies, dataset diversity, or model architecture design [10].

## 3. Methodology

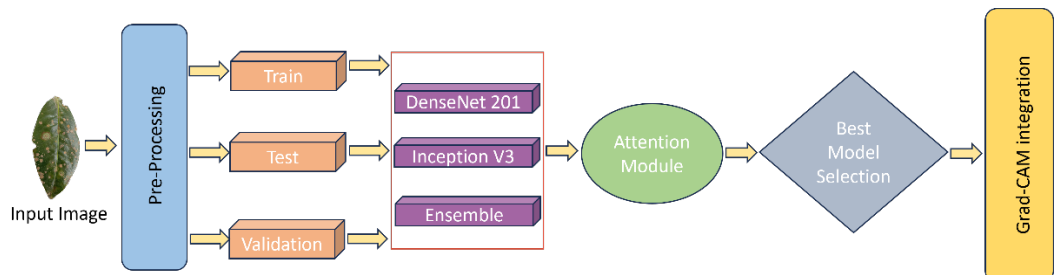

**Fig. 1.** Methodology

The methodology of research is presented in Fig 1.A novel dataset was created for this study. The dataset was then pre-processed prior training the models. Inception V3 and DenseNet201, these two pre-trained models were deployed. All the layers of the models were frozen except the last convolutional layer. A custom CNN model was placed on top of them. An attention module was placed in between the pre-trained models and the custom model. We utilized two different modules, Squeeze and Extraction (SE) along with Convolutional Block Attention Module (CBAM). Adversarial training was then introduced, leveraging the Fast Gradient Sign Method in order to make the model more robust. The optimal model was identified using multiple evaluation metrics. Explainable AI, namely Grad-CAM was integrated with the best model for better model interpretability.

### 3.1. Dataset Description and Preprocessing

We have created a novel dataset for this study [11]. There were 5278 photos across seven classes in the dataset. The seven classes are: Tea algal leaf spot, Brown Blight, Gray Blight, Helopeltis, Red spider, Green mirid bug and Healthy. A sample of our dataset is illustrated in Fig 2. The dataset was pre-processed using tensorflow's pre-processing library and 70 % of those were used for training, 20 % for validation and 10% for testing.

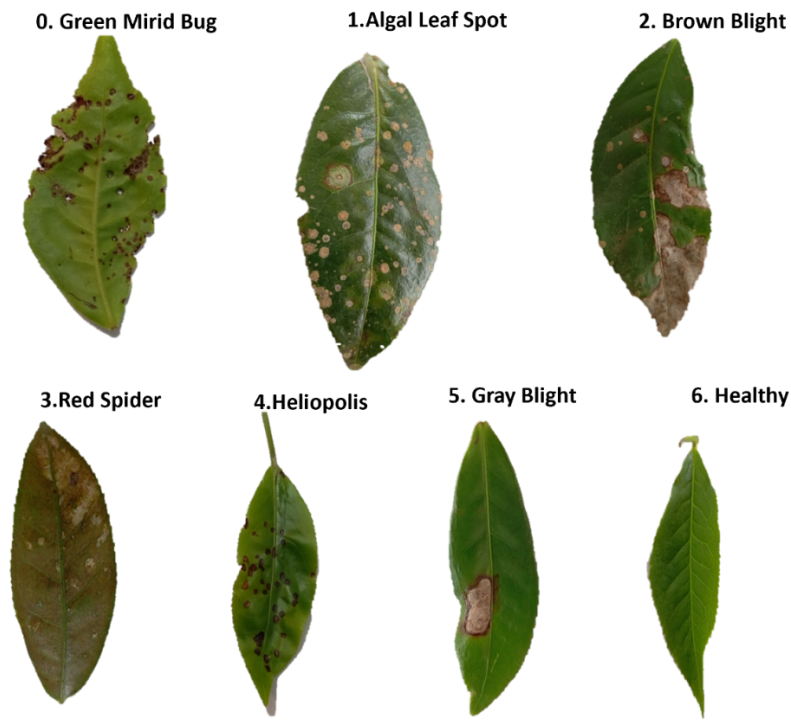

**Fig 2.** Sample inputs of the dataset

### 3.2. Pre-trained Models

A. DenseNet 201: DenseNet-201 is notable for its numerous interlayer connections, which facilitate feature reuse and address the vanishing gradient issue by linking each layer to every other layer in a feed-forward manner. With a total of 201 layers, including transition layers that regulate the number of feature maps.
B. Inception V3: Inception V3 is a highly efficient model for transfer learning, consisting of 48 layers. Its key components include auxiliary classifiers, reduction modules, asynchronous branching, 1x1 convolutions, and inception modules. The inception modules branch out in multiple directions, and the vanishing gradient problem is addressed by the auxiliary classifiers.
C. Ensemble Model: The ensemble technique merges multiple models to improve performance. In this study, we applied the soft voting method to create an ensemble model using DenseNet 201, Inception V3, and EfficientNet B4. Soft voting calculates the average probabilities for each class to combine predictions, and finally the prediction is made based on weighted probability.

### 3.3. Attention Module

A. Squeeze and Extraction (SE) block: The SE Block achieves feature learning by adaptive channel-wise feature response recalibration. It first compresses spatial information via Global Average Pooling (GAP) and then channel excitement via two Dense layers through a gating mechanism. The resulting attention weights weight input features, paying special attention to informative channels.
B. Convolutional Block Attention Module (CBAM): CBAM adjusts features in a channel sequence and spatial attention. Channel attention fuses GAP and Global Max Pooling (GMP) with an MLP, and spatial attention utilizes cross-channel pooling and a 7×7 Conv layer for focusing on relevant areas.

### 3.4. Custom Model

Our four-layer custom model was built and placed on top of Inception V3 and DenseNet201. The simplified architecture of our model is: Global Average Pooling (GAP)→ Dense Layer→ Final Dense Layer. Equation for the model as follows:

$$y = Softmax(W_2.Dropout(ReLU(W_1.GAP(F)+b_1))+b_2) \quad ....(1)$$

### 3.5. Hyperparameter Tuning

Hyperparameter tuning a key part of our study. We tweaked the pre-trained models and froze all their layers except for their last convolution layer. Hyper parameter tuning can greatly affect the model performance.

Table-I. Hyper Parameter Tuning

| Hyperparameter | Optimizer | Loss Function | Scheduler | Batch Size | Epochs | Patience |
|---|---|---|---|---|---|---|
| Value | Adam | Sparse Categorical Crossentropy | .0001 (multiply 1/10 every 5 epoch) | 32 | 50 | 10 |

### 3.6. Gradient-Weighted Class Activation Mapping

Grad-CAM is a extremely effective XAI method. It's foal is to visualize where a CNN is lookin when making a decision, especially in terms of which parts of the image are influencing the model's predictions. A forward and a backward pass is done in order to gradient calculation and to generate heatmaps. The equation for Grad-CAM is as follows:

For $k^{th}$ filter the feature map is $A^k$, $y^c$ is the score for class c, global average of gradient is,

$$a_c^k = \frac{1}{z}\sum_{i,j} \frac{\Delta y_c}{\Delta A^k} \ldots\ldots\ldots(6)$$

Class activation map $L_{\text{Grad-CAM}} = \text{ReLU}(\sum_k a_c^k \cdot A^k)\ldots\ldots(7)$

## 4. Implementation and Experimental Result

The performance of our model on test dataset for each method is shown in Table-II. We can see that the Ensemble model achieved the highest accuracy of 85.68% when used with CBAM. The performance of each model becomes better when SE attention module is used compare to using no attention module. The model performance becomes even better when CBAM is utilized. This table clearly highlights the effectiveness of CBAM and Ensemble techniques.

Table-II. Model Performance

| Models | SE block | CBAM block | No module |
|---|---|---|---|
| DenseNet201 | 76.27 | 78.31 | 73.92 |
| InceptionV3 | 80.46 | 82.93 | 77.28 |
| Ensemble model | 82.07 | 85.68 | 79.08 |

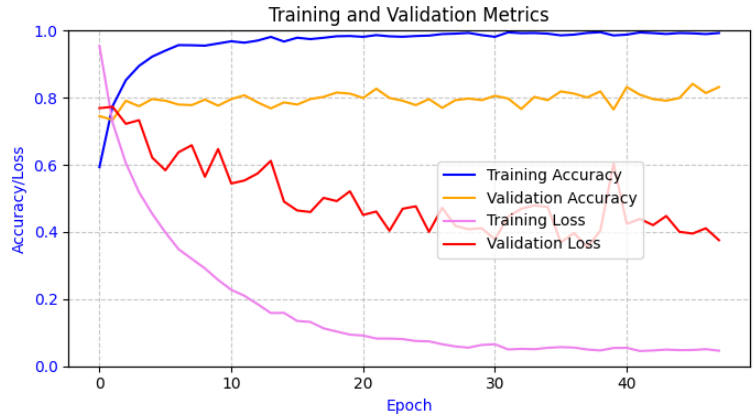

**Fig 3.** Accuracy and Loss of Ensemble model

Fig. 3 shows the performance of the ensemble model. We can see that the model performance is stable throughout the validation phase. There is no erratic behavior shown in the graph.

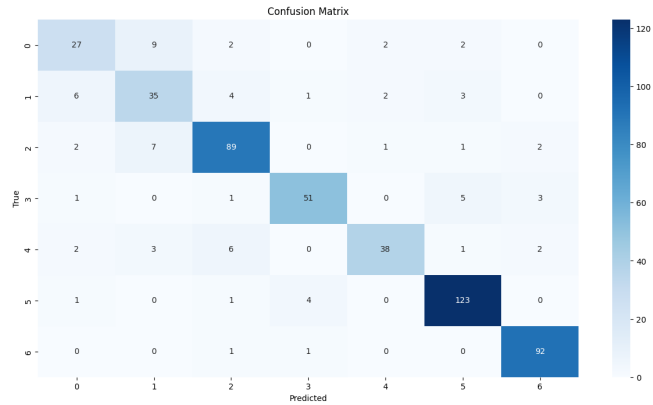

**Fig 4.** Confusion Matrix for the Ensemble Model

Confusion matrix for the same model is presented in Fig 4. The model performance is almost same for most classes. We also don't see a high number of misclassifications between any two specific classes that would indicate model is having trouble distinguishing those two classes which indicates that the model is capable of separating features of each class quite well. This is further proven by Fig 5 which illustrates the Receiver-operating characteristic curve (ROC) of Ensemble Model. The area under the curve (AUC) for each class is quite high. More than 95% for each class. Since, Ensemble model was our best model we integrate Grad-CAM with the Ensemble model. The heatmap presented in Fig 6 clearly highlights the affected area. The area marked in dark red is the area mainly responsible for the model prediction. The value of the area decreases as the highlighted color moves from red to yellow to blue. Yellow regions are moderately responsible while blue areas aren't responsible for the model prediction at all.

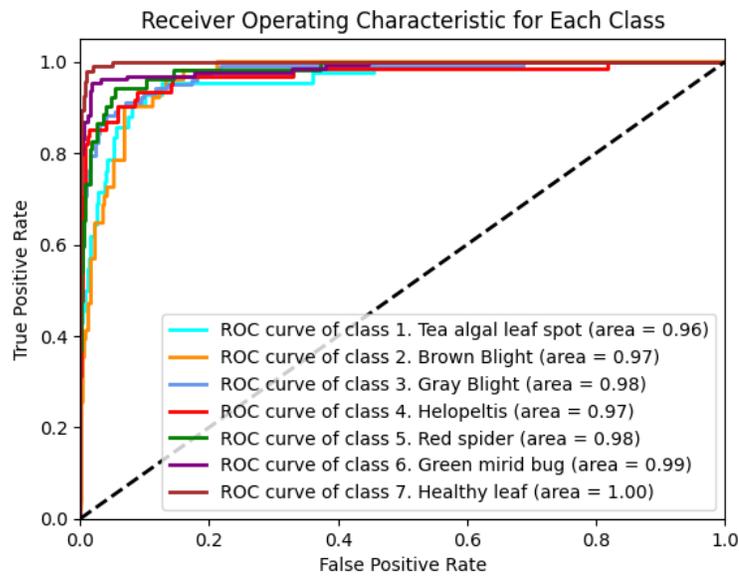

**Fig 5.** Receiver-operating characteristic curve (ROC) of Ensemble Model

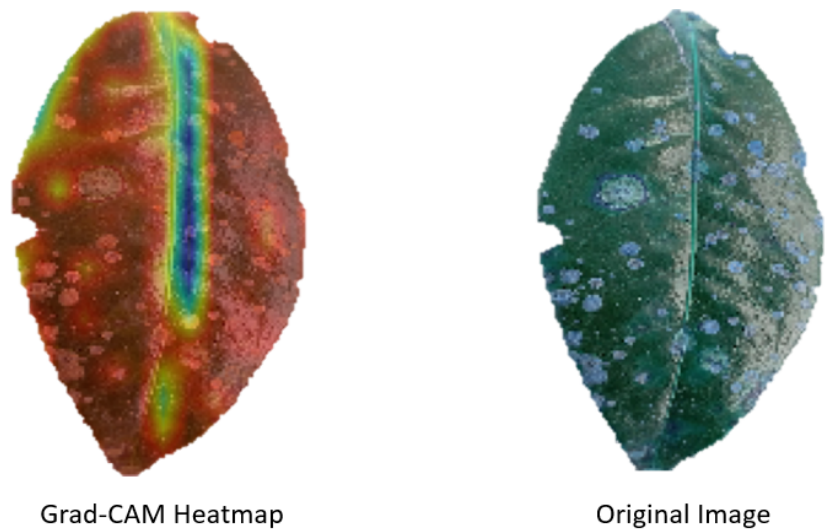

**Fig 6.** Heatmap generated by Grad-CAM

## 5. Conclusion

Tea leaf disease can be detected effectively by leveraging AI. In this study a dataset with 5278 images were utilized. The dataset was processed prior training. To apply transfer learning pre-trained models were deployed. All the layers of those models were frozen except their last convolutional block. To enhance model performance attention module and ensemble technique were introduced. The combination of both these techniques resulted in highest accuracy. The Ensemble model achieved the highest accuracy of 85.68% highlighting the effectiveness of ensemble technique and attention modules. In future works, the dataset can be expanded and models can be fine-tuned further for better result.

6. U. Hairah, A. Septiarini, N. Puspitasari, A. Tejawati, H. Hamdan, and S. E. Priyatna, "Classification of tea leaf disease using convolutional neural network approach," International Journal of Electrical and Computer Engineering, Jun. 2024, doi: 10.11591/ijece.v14i3.pp3287-3294.
7. A. Kaur, V. K. Kukreja, M. Manwal, N. Garg, and S. Hariharan, "Harnessing Deep Learning for Tea Tree Leaf Disease Management: A CNN-SVM Perspective," pp. 1–6, Dec. 2023, doi: 10.1109/punecon58714.2023.10450005.
8. Md. H. Rahman, I. Ahmad, P. H. Jon, A. Salam, and Md. F. Rabbi, "Automated detection of selected tea leaf diseases in Bangladesh with convolutional neural network," Dental science reports, vol. 14, no. 1, Jun. 2024, doi: 10.1038/s41598-024-62058-3.
9. Q. Heng, S. Yu, and Y. Zhang, "A new AI-based approach for automatic identification of tea leaf disease using deep neural network based on hybrid pooling," Heliyon, vol. 10, Feb. 2024.
10. G. Kaur, N. Sharma, and R. Gupta, "Tea Leaf Disease Detection using Deep Learning Convolutional Neural Network Model," pp. 1–6, Oct. 2023, doi: 10.1109/easct59475.2023.10393773.
11. Alam, B M Shahria; Ahammed, Fahad; Kibria, Golam; Noor, Mohammad Tahmid; Shikdar, Omar Faruq; Mahazabin, Kazi Isat; Ali, Md Nawab Yousuf (2025), "teaLeafBD", Mendeley Data, V4, doi: 10.17632/744vznw5k2.4